\documentclass[runningheads]{llncs}
\usepackage{amsmath,amssymb} 
\usepackage{color}
\usepackage[width=122mm,left=12mm,paperwidth=146mm,height=193mm,top=12mm,paperheight=217mm]{geometry}

\usepackage{graphicx}

\usepackage[utf8]{inputenc}
\usepackage{array}
\usepackage{makecell}
\usepackage{hyperref}

\usepackage{float}
\usepackage[title]{appendix}
\usepackage{breqn}
\usepackage{multirow}
\usepackage{hhline}
\usepackage{ltablex}
\usepackage{siunitx}
\usepackage{caption}
\usepackage{booktabs}
\usepackage{mathtools}
\usepackage{arydshln}
\DeclarePairedDelimiter{\norm}{\lVert}{\rVert}

\usepackage{float}
\usepackage{wrapfig}
\graphicspath{{images/}}

\newcommand*\samethanks[1][\value{footnote}]{\footnotemark[#1]}

\begin{document}
\pagestyle{headings}
\mainmatter

\title{Lip Movements Generation at a Glance} 



\titlerunning{Lip Movements Generation at a Glance}
\authorrunning{L. Chen, Z. Li, R. K. Maddox, Z. Duan, and C. Xu}

\author{Lele Chen$^1$\thanks{Equal contribution. This work was done when Zhiheng Li was a visiting student at University of Rochester.}, Zhiheng Li$^2$\samethanks, Ross K. Maddox$^{3}$, Zhiyao Duan$^4$, and Chenliang~Xu$^1$}


\institute{
$^1$Department of Computer Science, University of Rochester\\
$^2$School of Computer Science, Wuhan University\\ 
$^3$Department of Biomedical Engineering, University of Rochester\\
$^4$Department of Electrical and Computer Engineering, University of Rochester\\
\texttt{\{lchen63,zli82,rmaddox\}@ur.rochester.edu}\\
\texttt{\{zhiyao.duan,chenliang.xu\}@rochester.edu}\\
}

\maketitle


\begin{abstract}
Cross-modality generation is an emerging topic that aims to synthesize data in one modality based on information in a different modality. In this paper, we consider a task of such: given an arbitrary audio speech and one lip image of arbitrary target identity, generate synthesized lip movements of the target identity saying the speech. To perform well in this task, it inevitably requires a model to not only consider the retention of target identity, photo-realistic of synthesized images, consistency and smoothness of lip images in a sequence, but more importantly, learn the correlations between audio speech and lip movements. To solve the collective problems, we explore the best modeling of the audio-visual correlations in building and training a lip-movement generator network. Specifically, we devise a method to fuse audio and image embeddings to generate multiple lip images at once and propose a novel correlation loss to synchronize lip changes and speech changes. Our final model utilizes a combination of four losses for a comprehensive consideration in generating lip movements; it is trained in an end-to-end fashion and is robust to lip shapes, view angles and different facial characteristics. Thoughtful experiments on three datasets ranging from lab-recorded to lips in-the-wild show that our model significantly outperforms other state-of-the-art methods extended to this task.

\end{abstract}

\section{Introduction}

\begin{figure}[t]
\centering
\includegraphics[width=0.85\linewidth]{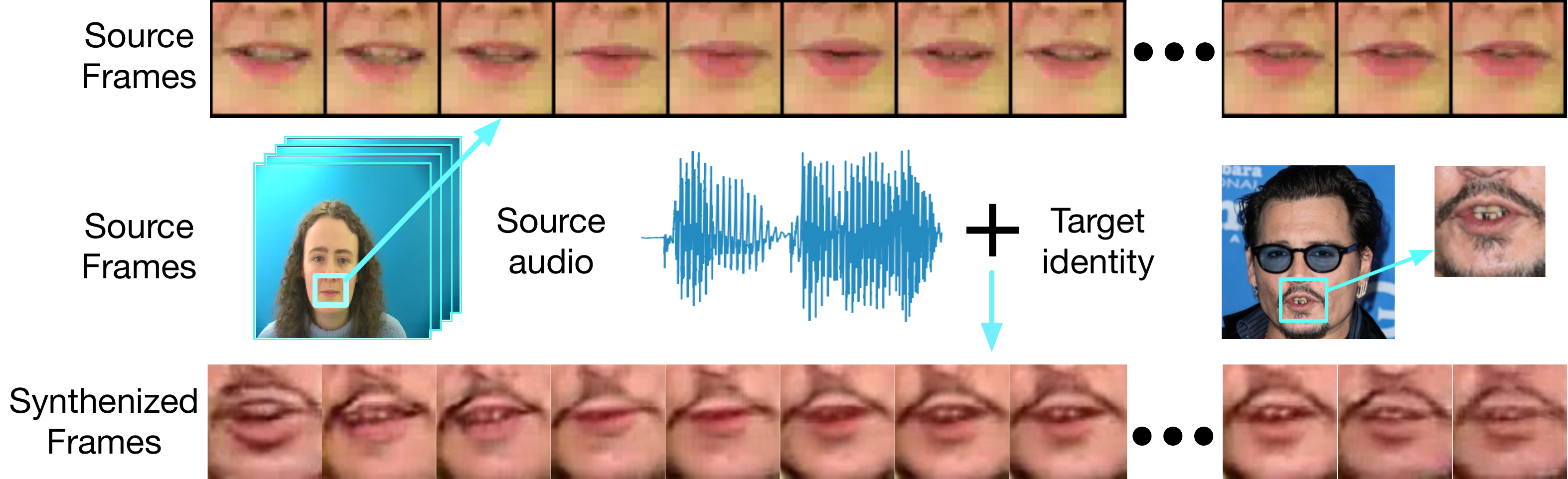}
\caption{The model takes an audio speech of the women and one lip image of the target identity, a male celebrity in this case, and synthesizes a video of the man's lip saying the same speech. The synthesized lip movements need to correspond to the speech audio and also maintain the target identity, video smoothness and sharpness.}
\vspace{-5mm}
\label{fig:example}
\end{figure}

Cross-modality generation has become an important and emerging topic of computer vision and its broader AI communities, where examples are beyond the most prominent image/video-to-text~\cite{DaXuDoCVPR2013,KuPrDhCVPR2011} and can be found in video-to-sound~\cite{OwIsMcCVPR2016}, text-to-image~\cite{ReedAYLSL16}, and even sound-to-image~\cite{DBLP:conf/mm/ChenSDX17}. This paper considers a task: given an arbitrary audio speech and one lip image of arbitrary target identity, generate synthesized lip movements of the target identity saying the speech. Notice that the speech does not have to be spoken by the target identity, and neither the speech nor the image of target identity is required to be appeared in the training set (see Fig.~\ref{fig:example}). Solving this task is crucial to many applications, e.g., enhancing speech comprehension while preserving privacy or assistive devices for hearing impaired people.


Lip movements generation has been traditionally solved as a sub-problem in synthesizing a talking face from speech audio of a target identity~\cite{DBLP:conf/icassp/FanWSX15,DBLP:journals/cgf/GarridoVSSVPT15,Charles2016,DBLP:journals/tog/SuwajanakornSK17}. For example, Bo et al.~\cite{DBLP:conf/icassp/FanWSX15} restitch the lower half of the face via a bi-directional LSTM to re-dub a target video from a different audio source. Their model selects a target mouth region from a dictionary of saved target frames. More recently, Suwajanakorn et al.~\cite{DBLP:journals/tog/SuwajanakornSK17} generate synthesized taking face of President Obama with accurate lip synchronization, given his speech audio. They first use an LSTM model trained on many hours of his weekly address footage to generate mouth landmarks, then retrieve mapped texture and apply complicated post-processing to sharpen the generated video. However, one common problem for these many methods is that they retrieve rather than generating images and thus, require a sizable amount of video frames of the target identity to choose from, whereas our method generates lip movements from a single image of the target identity, i.e., at a glance. 


The only work we are aware of that addresses the same task as ours is Chung et al.~\cite{DBLP:journals/corr/ChungJZ17}. They propose an image generator network with skip-connections, and optimize the reconstruction loss between synthesized images and real images. Each time, their model generates one image from 0.35-second audio. Although their video generated image-by-image and enhanced by post-processing looks fine, they have essentially bypassed the harder questions concerning the consistency and smoothness of images in a sequence, as well as the temporal correlations of audio speech and lip movements in a video.


To overcome the above limitations, we propose a novel method that takes speech audio and a lip image of the target identity as input, and generates multiple lip images (16 frames) in a video depicting the corresponding lip movements (see Fig.~\ref{fig:example}). Observing that speech is highly correlated with lip movements even across identities, a concept grounds lip reading~\cite{DBLP:journals/corr/AssaelSWF16,DBLP:conf/accv/ChungZ16}, the core of our paper is to explore the best modeling of such correlations in building and training a lip movement generator network. To achieve this goal, we devise a method to fuse time-series audio embedding and identity image embedding in generating multiple lip images, and propose a novel audio-visual correlation loss to synchronize lip changes and speech changes in a video. Our final model utilizes a combination of four losses including the proposed audio-visual correlation loss, a novel three-stream adversarial learning loss to guide a discriminator to judge both image quality and motion quality, a feature-space loss to minimize perceptual-level differences, and a reconstruction loss to minimize pixel-level differences, for a comprehensive consideration of lip movements generation. The whole system is trained in an end-to-end fashion and is robust to lip shapes, view angles, and different facial characteristics (e.g., beard v.s. no beard).


We evaluate our model along with its variants on three datasets: The GRID audiovisual sentence corpus (GRID)~\cite{Cooke2006}, Linguistic Data Consortium (LDC)~\cite{richie2009audiovisual} and Lip Reading in the Wild (LRW)~\cite{DBLP:conf/accv/ChungZ16}. To measure the quantitative accuracy of lip movements, we propose a novel metric that evaluates the detected landmark distance of synthesized lips to ground-truth lips. In addition, we use a cohort of three metrics, Peak Signal to Noise Ratio (PSNR), Structure Similarity Index Measure (SSIM)~\cite{DBLP:journals/tip/WangBSS04}, and perceptual-based no-reference objective image sharpness metric (CPBD)~\cite{DBLP:journals/tip/NarvekarK11}, to measure the quality of synthesized lip images, e.g., image sharpness. We compare our model with Chung et al.~\cite{DBLP:journals/corr/ChungJZ17} and an extended version of the state-of-the-art video Generative Adversarial Network (GAN) model~\cite{DBLP:journals/corr/VondrickPT16} to our task. Experimental results show that our model outperforms them significantly on all three datasets (see "Full model" in Tab.~\ref{tab:main_tb}). Furthermore, we also show real-world novel examples of synthesized lip movements of celebrities, who are not in our dataset. Our demo video can be found in \url{https://youtu.be/7IX_sIL5v0c}. 


Our paper marks three contributions. First, to the best of our knowledge, we are the first to consider the correlations among speech and lip movements in generating multiple lip images at a glance. Second, we explore various models and loss functions in building and training a lip movement generator network. Third, we quantify the evaluation metrics and our final model achieves significant improvement over state-of-the-art methods extended to this task on three datasets ranging from lab-recorded to lips in-the-wild.


The rest of our paper is organized as follows: Sec.~\ref{sec:related_work} contains related work, Sec.~\ref{sec:network} depicts our generator network for lip movements and introduces our audio-visual correlation loss, Sec.~\ref{sec:fullmodal} details our full model and training procedure, Sec.~\ref{sec:experiments} shows experimental results and Sec.~\ref{sec:conclusion} concludes the paper and points out directions for future work.


\section{Related Work}
\label{sec:related_work}

We have briefly surveyed work in lip movement generation in the Introduction section. Here, we discuss related work of each techniques used in our model. 

A related but different task to ours is lip reading, where it also tackles the cross-modality generation problem. \cite{DBLP:journals/corr/AssaelSWF16,DBLP:conf/accv/ChungZ16} use the correlation between lip movement and the sentences/words to interpret the audio information from the visual information. Rasiwasia et al.~\cite{DBLP:conf/mm/RasiwasiaPCDLLV10} use Canonical Correlation Analysis (CCA)~\cite{hotelling1936relations} subspace learning to learn two intermediate feature spaces for two modalities where they do correlation on the projected features. Cutler and Davis~\cite{DBLP:conf/icmcs/CutlerD00} use Time Delay Neural Network~\cite{DBLP:journals/tsp/WaibelHHSL89} (TDNN) to extract temporal invariant audio features and visual features. These works have inspired us to model correlations between speech audio and lip movements in generating videos. 

Audio variations and lip movements are not always synchronized in the production of human speech; lips often move before the audio signal is produced~\cite{10.1371/journal.pcbi.1000436}. Such delay between audio and visual needs to be considered when designing a model. Suwajanakorn et al.~\cite{DBLP:journals/tog/SuwajanakornSK17} apply a time-delayed RNN without outputting value in the first few RNN cells. Therefore, the output is shifted accordingly to the delayed steps. However, such delay is empirically fixed by hand and thus, it is hard to determine the amount of delay for videos in-the-wild. We follow~\cite{DBLP:journals/tsp/WaibelHHSL89} to extract  features with a large receptive field along temporal dimension, but use a convolutional network instead of TDNN that leads to a simpler design.

Adversarial training~\cite{DBLP:journals/corr/GoodfellowPMXWOCB14} is recently introduced as a novel and effective way to train generative models. Researchers find that by conditioning the model on additional information, it is possible to direct the data generation process~\cite{DBLP:journals/corr/MirzaO14,DBLP:conf/nips/ChenCDHSSA16,DBLP:conf/icml/OdenaOS17}. Furthermore, GAN has shown its ability to bridge the gap between different modalities and produce useful joint representations. We also use GAN loss in our training but we show that combining it with other losses leads to better results.


 \section{Lip-Movement Generator Network}
\label{sec:network}

\begin{figure*}[t]
\centering\includegraphics[width=0.8\linewidth]{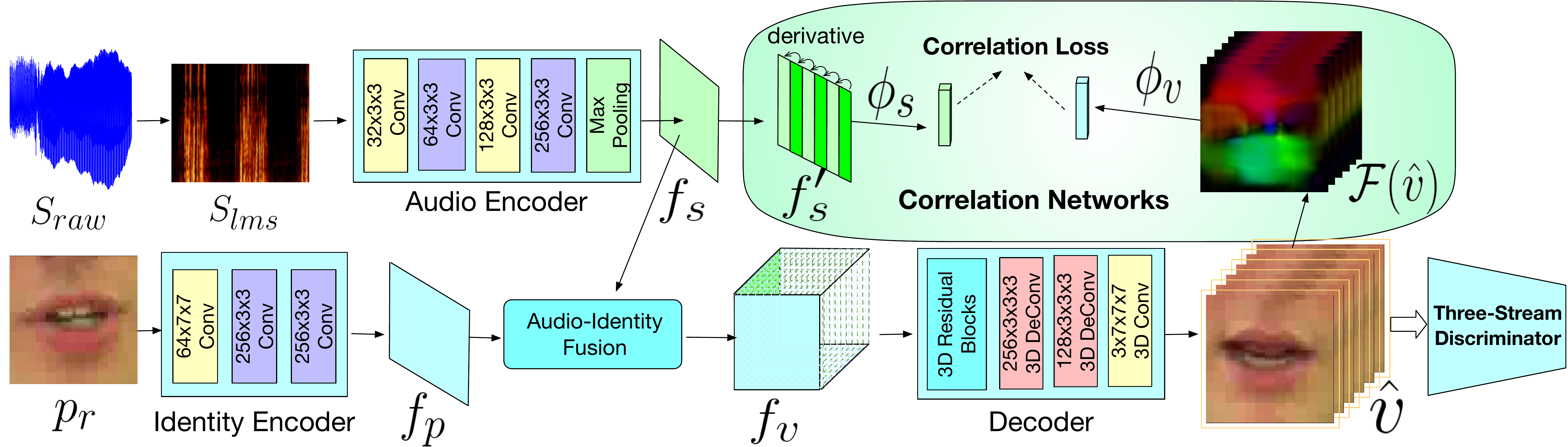}
\caption{Full model illustration. Audio encoder and identity encoder extracts and fuses audio and visual embeddings. Audio-Identity fusion network fuses features from two modalities. Decoder expands fused feature to synthesized video. Correlation Networks are in charge of strengthening the audio-visual mapping. Three-Stream discriminator is responsible for distinguishing generated video and real video.
}
\label{fig:generator}
\vspace{-3mm}
\end{figure*}


The overall data flow of our lip-movement generator network is depicted in Fig.~\ref{fig:generator}. \textbf{In this paper, we omit channel dimension of all tensors for simple illustration}. Recall that the input to our network are a speech audio and one single image of the target identity, and the output of our network are synthesized lip images of the target identity saying that audio. The synthesized lip movements need to correspond to the speech audio, maintain the target identity, ensure the video smoothness, and be photo-realistic.



\subsection{Audio-Identity Fusion and Generation}
\label{ssec:fusion}
First, we encode the two-stream input information. For audio stream, the raw audio waveform, denoted as $S_{raw}$, is first transformed into log-mel spectrogram (see detail in Sec.~\ref{ssec:exp:data}), denoted as $S_{lms}$, then encoded by an audio encoder network into audio features $f_s \in \mathbb{R}^{T \times F}$, where $T$ and $F$ denote the number of time frames and frequency channels. For visual stream, an input identity image, denoted as $p_r$, is encoded by an identity encoder network. The network outputs image features $f_p \in \mathbb{R}^{H \times W}$, where $H$ and $W$ denote the height, width of the output image features.

We fuse audio features $f_s$ and visual features $f_p$ together, whose output, the synthesized video feature $f_v$, will be expanded by several residual blocks and 3D deconvolution operations to generate synthesized video $\hat{v}$.
In order to make sure the synthesized clip is based on the target person and also captures the time-variation of speech, we investigate an effective way to fuse $f_s$ and $f_p$ to get $f_v$ for generating a video. Here, the challenge is that the feature maps exist in different modalities, e.g., audio, visual, and audio-visual, and reside in different feature spaces, e.g., time-frequency, space, and space-time. 
\begin{wrapfigure}{r}{0.5\textwidth}
\vspace{-5mm}
\includegraphics[width=0.49\textwidth]{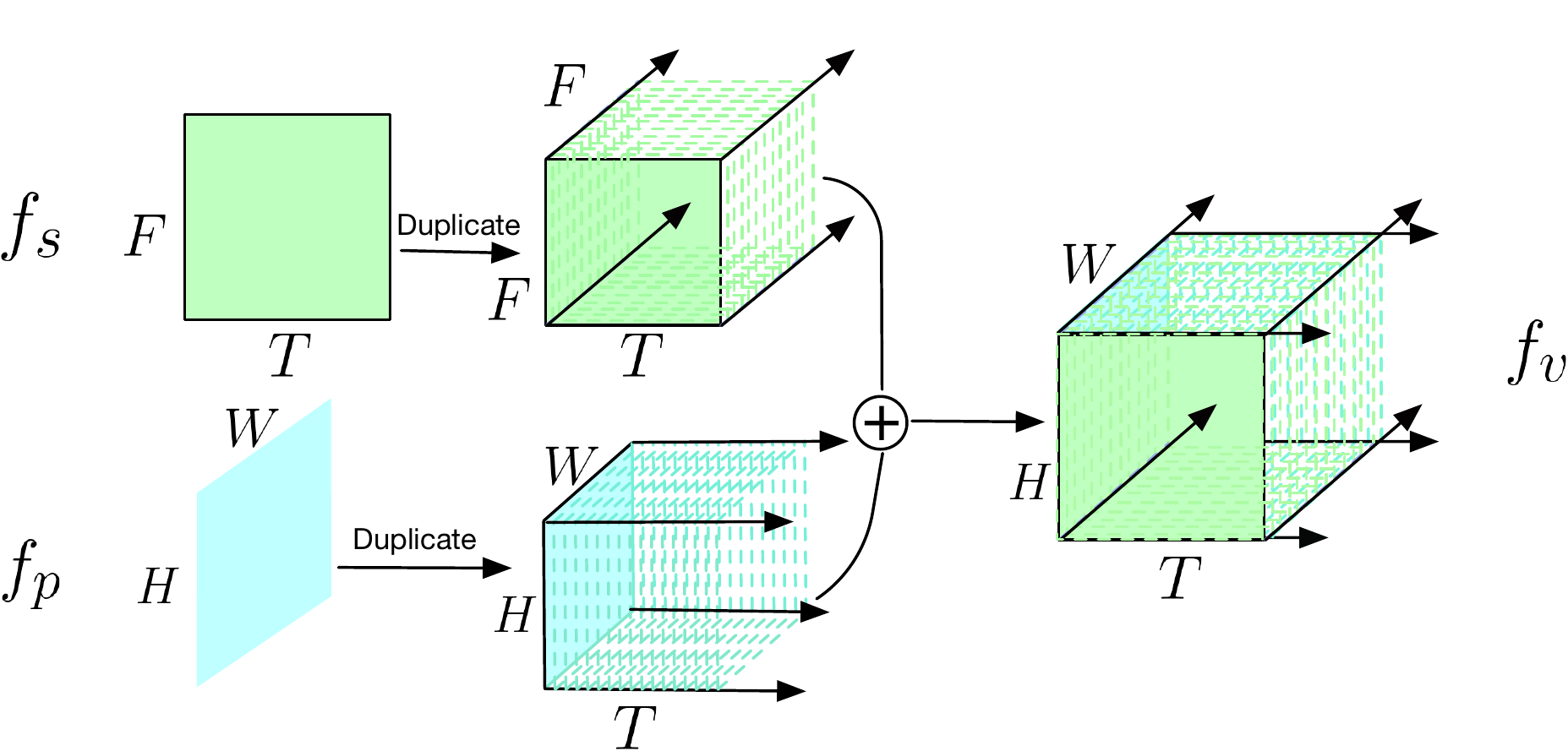}
\caption{Audio-Identity fusion. Transfer audio time-frequency features and image spatial features to video spatial-temporal features.}
\label{fig:spatial_temporal}
\vspace{-5mm}
\end{wrapfigure}


Our fusion method is based on duplication and concatenation. This process is depicted in Fig.~\ref{fig:spatial_temporal}. For each audio feature, we duplicate that feature along frequency dimension in each time step, i.e., from the size of $T \times F$ to the size of $T \times F  \times  F $. Image feature, which can be viewed as a template for video representation, is copied T times, i.e., from $H \times W$ to a new size $T \times H \times W$. We set $H=W=F$ in this method. Then, two kinds of duplicated features are concatenated along channel dimension.






\subsection{Audio-Visual Derivative Correlation Loss}
\label{sec:correlation}

We believe that the acoustic information of audio speech is correlated with lip movements even across identities because of their shared high-level representation. Besides, we also regard that variation along temporal axis between two modalities are more likely to be correlated. In other words, compared with acoustic feature and visual feature of lip shape themselves, the variation of audio feature (e.g. the voice raising to a higher pitch) and variation of visual feature (e.g. mouth opening) have a higher likelihood to be correlated. 
Therefore, we propose a method to optimize the correlations of the two modalities in their feature spaces. We use $f_s'$ in size of $(T-1) \times F$, the derivative of audio feature $f_s$ (with size of $T \times F$) between consecutive frames in temporal dimension, to represent the changes in speech. It goes through an audio derivative encoder network $\phi_s$, and thus, we have audio derivative feature $\phi_s(f_s')$. Similarly, we use $\mathcal{F}(v)$ to represent optical flows of each consecutive frames in a video $v$, where $\mathcal{F}$ is an optical flow estimation algorithm. It goes through an optical flow encoder network $\phi_v$, and thus, we have $\phi_v(\mathcal{F}(v))$ to depict the visual variations of lip movements in the feature space. We use cosine similarity loss to maximize the correlation between audio derivative feature and visual derivative feature:
	\begin{align}
\label{eq:corr}
\ell_{corr}  =  1 - \frac{\phi_s(f_s') \cdot \phi_v(\mathcal{F}(v))}{\norm{\phi_s(f_s')}_2\cdot\norm{\phi_v(\mathcal{F}(v))}_2} 
\enspace. 
\end{align}
%
Here, the optical flow algorithm applied to the synthesized frames needs to be differentiable for back-propagation~\cite{rumelhart1988learning}. In our implementation, we add a small number ($\epsilon = 10^{-8}$) to the denominator to avoid division by zero. In order to avoid trivial solution when $\phi_s$ and $\phi_v$ are learned to predict constant outputs $\phi_s(f_x')$ and $\phi_v(\mathcal{F}(v))$ which are perfectly correlated and the $\ell_{corr}$ will go to 0, we combine other losses during the training process (see Eq.~\ref{eq:full_loss}).



\begin{figure}[t]
\begin{center}
\includegraphics[width=0.75\linewidth]{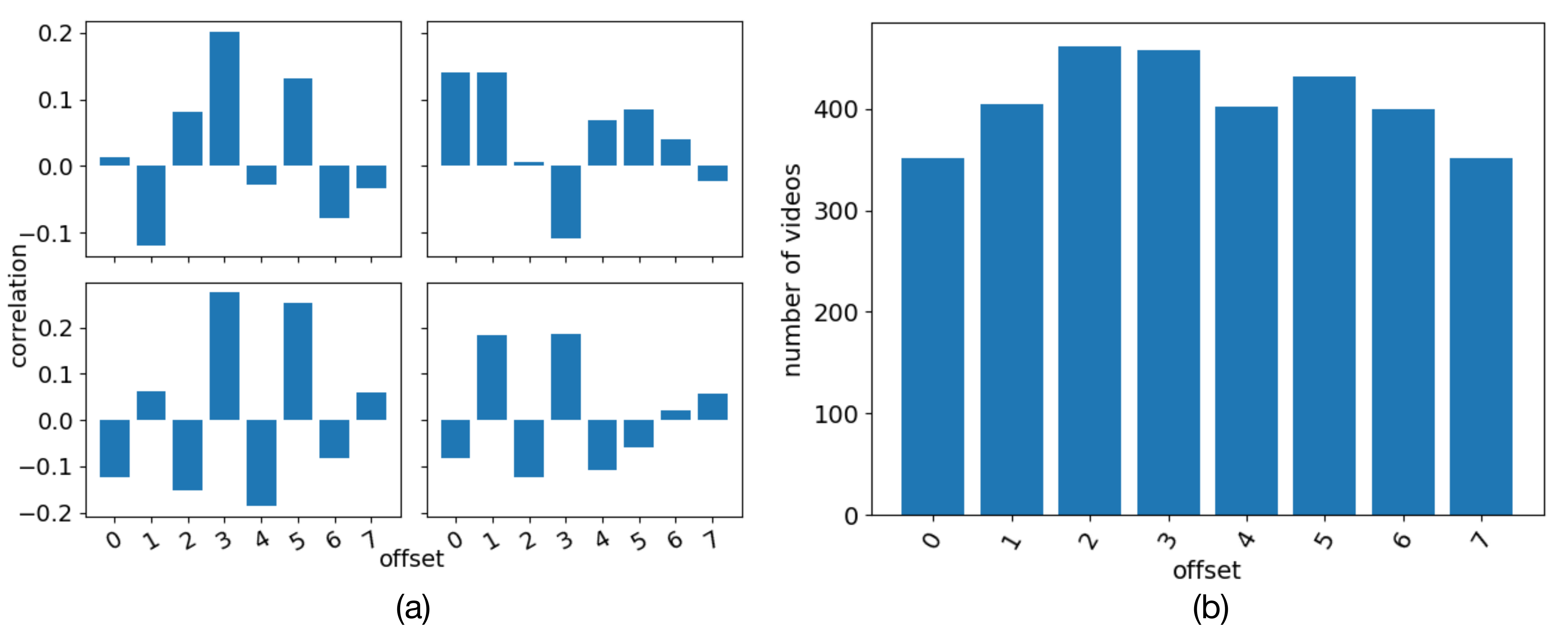}
\end{center}
\vspace{-3mm}
\caption{(a): Correlation coefficients with different offsets of four example videos. (b): Number of videos of different offsets with which the video has the maximum correlation coefficient. X-axes of both (a) and (b) stands for time steps of flow field shifted forward.
}
\label{fig:corr_analysis}
\vspace{-3mm}
\end{figure}




\noindent \textbf{Correlation Networks.} \quad The audio and visual information are not perfectly aligned in time. Usually, lip shape forms earlier than sound. For instance, when we say word `bed', upper and lower lips meet before speaking the word~\cite{10.1371/journal.pcbi.1000436}. If such delay problem exists, aforementioned correlation loss, assuming audio-visual information are perfectly aligned, may not work. We verify the delayed correspondence problem between audio and visual information by designing a case study on 3260 videos randomly sampled from the GRID dataset. The solution for the delayed correspondence problem is given in the next paragraph. In the case study, for each 75-frame video $v$, we calculate the mean values of each 74 derivatives of audio $s_{lms}$ and mean values of each 74 optical flow fields $\phi_v(\mathcal{F}(v))$. With respect to each video, we shift mean values of optical flows forward along time at different offsets (0 to 7 in our case study) and calculate Pearson correlation coefficients of those two parts. Results of four videos, calculated by aforementioned procedures, are shown in Fig.\ref{fig:corr_analysis}(a). Finally, we count the number of videos in different offsets at which the video has the largest correlation coefficient, as shown in Fig.~\ref{fig:corr_analysis}(b). Figure \ref{fig:corr_analysis} shows that different videos prefer different offsets to output the maximum correlation coefficient, which indicates that fixing a constant offset of all audio-visual inputs would not solve the problem of correlation with inconsistent delays among all videos in a dataset.

To mitigate such delayed correlation problem, we design correlation networks (as shown in Fig.~\ref{fig:generator}) containing an audio derivative encoder $\phi_s$ and an optical flow encoder $\phi_v$ to extract features used for calculating the correlation loss in Eq.~\ref{eq:corr}.
These networks reduce the feature size but retain the temporal length simultaneously. The sizes of the two outputs are matched for calculating the correlation loss. We use 3D CNNs to implement these networks, which are also helpful to mitigate the fixed offset problem happens in previous works \cite{DBLP:journals/tog/SuwajanakornSK17}. Both $\phi_s$ and $\phi_v$ output features with large receptive fields (9 for $\phi_s(f_s')$ and 13 for $\phi_v(\mathcal{F}(v))$), which consider the audio-visual correlation in large temporal dimension. Compared with time-delayed RNN proposed in~\cite{DBLP:journals/tog/SuwajanakornSK17}, CNN can learn delay from the dataset rather than set it as a hyperparameter. Besides, CNN architecture benefits from its weight sharing property leading to a simpler and smaller design than TDNN~\cite{DBLP:journals/tsp/WaibelHHSL89}.


\section{Full Model and Training}
\label{sec:fullmodal}



Without loss of generality, we use pairs of lip movement video and speech audio $\{(v^j, s^j)\}$ in the training stage, where $v^j$ represents the $j$th video in our dataset and $s^{j}$ represents the corresponding speech audio. We omit the superscript $j$ when it is not necessary for the discussion of one sample. We use $p_r$ to denote one lip image of the target speaker, which can provide the initial texture information. During training, we train over $(v,s)$ in the training set and sample $p_r$ to be one frame randomly selected from the raw video where $v^j$ is sampled from to ensure that $v$ and $p_r$ contain the same identity.
Therefore, the system is robust to the lip shape of  the identity $p_{r}$. The objective of training is to generate a realistic video $\hat{v}$ that resembles $v$. For testing, the speech $s$ and identity image $p_r$ can be any speech and any lip image (even out of the dataset we used in training). Next, we present the full model in the context of training. 

Our full model (see Fig.~\ref{fig:generator})
is end-to-end trainable and is optimized according to the following objective function: 
\begin{align}
\label{eq:full_loss}
\mathcal{L} =  \ell_{corr}  + \lambda_{1} \ell_{pix} + \lambda_{2} \ell_{perc} + \lambda_{3} \ell_{gen}
\enspace,
\end{align}
where $\lambda_{1}$, $\lambda_{2}$ and $\lambda_{3}$ are coefficients of different loss terms. We set them as 0.5, 1.0, 1.0 respectively in this paper. The intuitions behind the four losses are as follows:
\begin{itemize}
\setlength\itemsep{0em}
\item \textbf{$\ell_{corr}$}: \quad Correlation loss, illustrated in Sec.~\ref{sec:correlation}, is introduced to ensure the correlation between audio and visual information.
\item \textbf{$\ell_{pix}$}:\quad Pixel-level reconstruction loss, defined as $\ell_{pix} (\hat{v},v) = \norm{v - \hat{v}}$, which aims to make the model sensitive to speaker's appearance, i.e., retain the identity texture. However, we find that using it alone will reduce the sharpness of the synthesized video frames.
\item \textbf{$\ell_{perc}$}: \quad Perceptual loss, which is originally proposed by~\cite{DBLP:conf/eccv/JohnsonAF16} as a method used in image style transfer and super-resolution. It utilizes high-level features to compare generated images and ground-truth images, resulting in better sharpness of the synthesized image. We adapt this perceptual loss and detail it in Sec.~\ref{subsec:perceptual}. 
\item \textbf{$\ell_{gen}$}: \quad Adversarial loss allows our model to generate overall realistic looking images and is defined as: $\ell_{gen}= - \log {D([s^j,\hat{v}^j])}$, where $D$ is a discriminator network. We describe the detail of our proposed stream-stream GAN discriminator in Sec.~\ref{ssec:discriminator}.
\end{itemize}



\subsection{Autoencoder and Perceptual Loss}
\label{subsec:perceptual}

In order to avoid over-smoothed phenomenon of synthesized video frames $\hat{v}$, we adapt perceptual loss proposed by Johnson et al.~\cite{DBLP:conf/eccv/JohnsonAF16}, which reflects perceptual-level similarity of images. The perceptual loss is defined as: 
\begin{align}
\label{eq:perceptual}
\ell_{perc} (\hat{v},v) = 
\norm{\varphi(v) - \varphi(\hat{v})}_{2}^2
\enspace,
\end{align}
where $\varphi$ is a feature extraction network. We train an autoencoder to reconstruct video clips. To let the network be more sensitive to structure features, we apply six residual blocks after the convolution layers. We train the autoencoder from scratch, then fix the weights and use its encoder part as $\varphi$ to calculate perceptual loss for training the full model.
\subsection{Three-Stream GAN Discriminator}
\label{ssec:discriminator}
\begin{wrapfigure}{r}{0.5\textwidth}
\vspace{-5mm}
\includegraphics[width=0.49\textwidth]{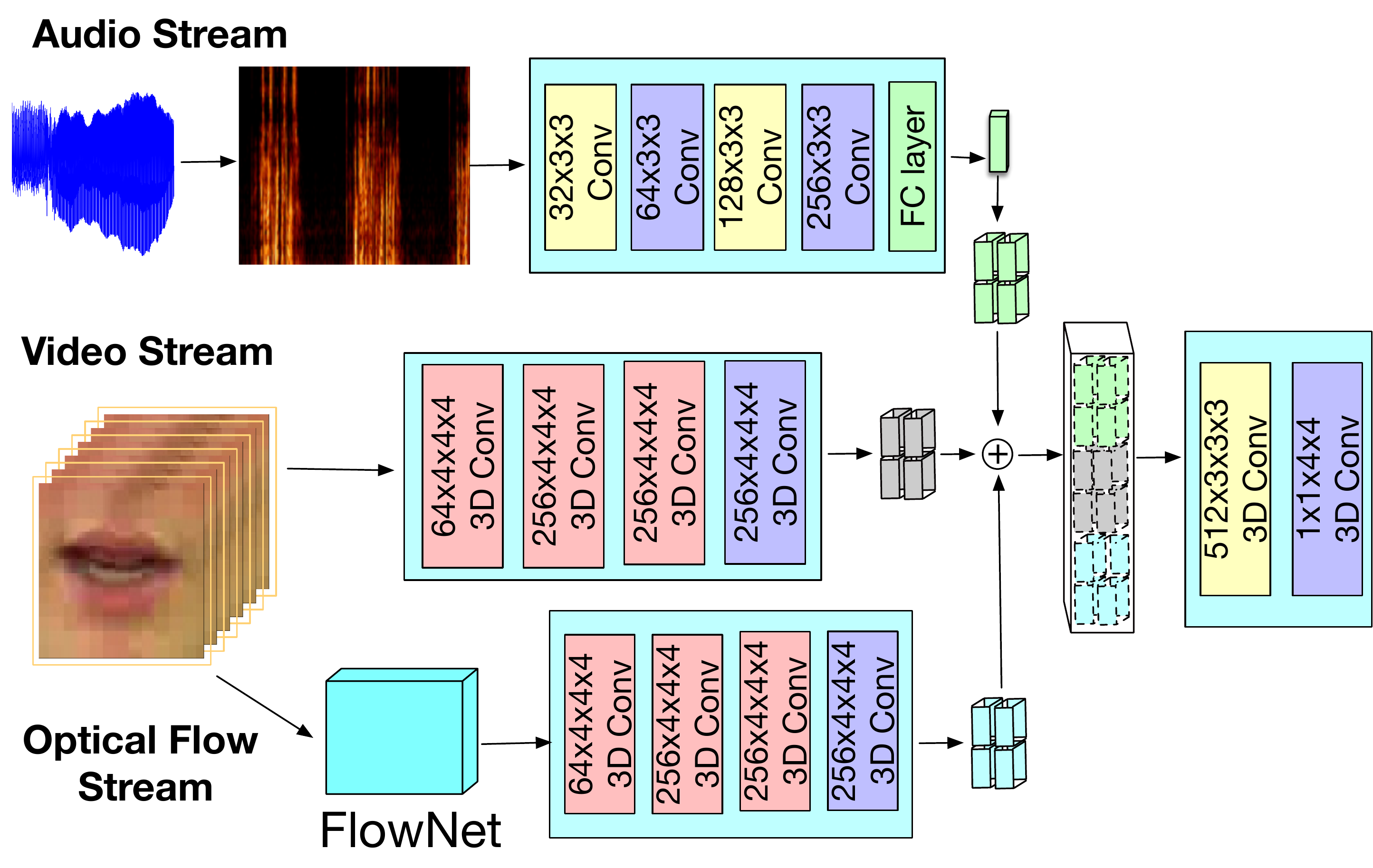}
\caption{Three-stream GAN discriminator illustration.}
\label{fig:discriminator}
\vspace{-5mm}
\end{wrapfigure}
The GAN discriminator in~\cite{DBLP:journals/corr/VondrickPT16} for synthesizing video considers the motion changes implicitly by 3D convolution. In order to generate sharp and smooth changing video frames, we propose a three-stream discriminator network (see fig.~\ref{fig:discriminator}) to distinguish the synthesized video ($\hat{v}^j$) from real video ($v^j$) that not only considers motion explicitly and but also conditions on the input speech signal. The input to the discriminator is a video clip with the corresponding audio. We have the following three streams. For audio stream (also used in our generator), we first convert the raw audio to log-mel spectrogram, then use four convolutional layers followed by a fully-connected layer to get a 1D vector. We duplicate it to match features from other streams. For video stream, we use four 3D CNN layers to extract video features. In addition, we include an optical flow stream that attends to motion changes explicitly. We fine-tune the FlowNet~\cite{DBLP:conf/iccv/DosovitskiyFIHH15}, which is pre-trained on FlyingChairs dataset, to extract optical flows, then apply four 3D CNN layers to extract features.

Finally, we concatenate the three-stream features in channel dimension and let them go through two convolutional layers to output the discriminator probability. We adapt mismatch strategy~\cite{ReedAYLSL16} to make sure that our discriminator is also sensitive to mismatched audio and visual information. Therefore, the discriminator loss is defined as: 
\begin{align}
\ell_{dis}= &- \log{D([s^j,v^j])}   \nonumber \\
& -  \lambda_p \log { (1 - D([s^j,\hat{v}]))}  \nonumber \\
& - \lambda_u \log { (1 - D([s^j,{v}^k]))} , \quad k \neq j
\enspace,
\label{eq:dis}
\end{align}
where $v^k$ represents a mismatch real video. We set both $\lambda_p$ and $\lambda_u$ $0.5$ in our experiment. The performance of the optical flow stream is discussed in Sec.~\ref{ssec:ablation}


\section{Experiments}
\label{sec:experiments}


In this section, we first introduce datasets and experimental settings, and our adapted evaluation metrics. Then, we show ablation study and comparison to the state of the art. Finally, we demonstrate real-world novel examples. 

\subsection{Datasets and Settings}
\label{ssec:exp:data}

We present our experiments on GRID~\cite{Cooke2006}, LRW \cite{DBLP:conf/accv/ChungZ16} and LDC~\cite{richie2009audiovisual} datasets. We crop all the mouth regions as our focused region. There are 33 different speakers in GRID. Each speaker has 1000 short videos. The LRW dataset consists of 500 different words spoken by hundreds of different speakers. There are 14 speakers in the LDC dataset in which each speaker reads 238 different words and 166 different sentences. Videos in GRID and LDC are lab-recorded while videos in LRW are collected from news. The basic dataset information is shown in Tab.~\ref{tab:dataset}. Our data is composed by two parts: audio and image frames. The network can output different numbers of frames. In this work, we only consider generating 16 image frames. As the videos are sampled at 25 fps, the time span of the synthesized image frames is 0.64 seconds. We use sliding window approach (window size: 16 frames, overlap: 8 frames) to obtain training and testing video samples over raw videos.

\noindent \textbf{Audio:} \quad We extract audio from the video file with a sampling rate of 41.1 kHz. Each input audio is 0.64 seconds long (0.04 $\times$ 16). To encode audio, we first transform the raw audio waveform into the time-frequency domain by calculating the Log-amplitude Mel-frequency Spectrum (LMS). When we calculate the LMS, the number of samples between successive frames, the length of the FFT window, and the number of Mel bands are 512, 1024 and 128, respectively. This operation will convert a 0.64-sec raw audio to a 64 $\times$ 128 time-frequency representation.

\noindent \textbf{Images:} \quad First, we extract all image frames from videos. Then, we extract lip landmarks~\cite{DBLP:journals/jmlr/King09} and crop the image around the lip. Landmarks are only used for cropping and evaluation. We resize all of the cropped images to 64 $\times$ 64 and normalize all images (mean $=$ 0.5, std $=$ 0.5) along channel dimension. So, each 0.64-sec audio corresponds to a 16 $\times$ 3 $\times$ 64 $\times$ 64 RGB image sequence.

We adopt Adam optimizer during training and fixed learning rates of $10 \textsuperscript{-4}$ with weight decay of $5 \times 10 \textsuperscript{-4}$. We initialize all network layers according to the method described in~\cite{DBLP:conf/iccv/HeZRS15}. All models are trained and tested on a single NVIDIA GTX 1080Ti. During testing, generating one single frame costs $0.015$ seconds.

\begin{table}[t]
\begin{center}
\begin{tabular}{c c c c}
\toprule
\toprule
Dataset & GRID & LRW & LDC \\
\hline

Train & 211k ($37.5 h$) & 841k $(159.8 h)$ &36k ($6.4 h$) \\
\hline
Val. & 23k ($4.2 h$)  & N/A &4k ($0.7 h$) \\
\hline
Test& 7k ($1.3 h$) &40k ($7.8 h$) & 6.6k ($1.2 h$)\\
\bottomrule
\end{tabular}
\end{center}
\vspace{-1mm}
\caption{Dataset information. Validation set: known speakers but unseen sentences. Testing set: unseen speakers and unseen sentences.}
\label{tab:dataset}
\vspace{-5mm}
\end{table}

\subsection{Evaluation Metrics}
\label{sec:exp:metrics}

To evaluate the quality of the synthesized video frames, we compute Peak Signal to Noise Ratio (PSNR) and Structure Similarity Index Measure (SSIM)~\cite{DBLP:journals/tip/WangBSS04}. To evaluate the sharpness of the generated image frames, we compute the perceptual-based no-reference objective image sharpness metric (CPBD)~\cite{DBLP:journals/tip/NarvekarK11}.

As far as we know, no quantitative metric has been used to evaluate the accuracy of generated lip movements video. Therefore, to evaluate whether the synthesized video $\hat{v}$ corresponds to accurate lip movements based on the input audio, a new metric is proposed by calculating the Landmark Distance (LMD). We use Dlib~\cite{DBLP:journals/jmlr/King09}, a HOG-based facial landmarks detector, which is also widely used in lip-movement generation task and other related works\cite{DBLP:journals/tog/SuwajanakornSK17,Chung_2017_CVPR}, to detect lip landmarks on $\hat{v}$ and $v$, and mark them as $LF$ and $LR$, respectively. To eliminate the geometric difference, we calibrate the two mean points of lip landmarks in $LF$ and $LR$. Then, we calculate the Euclidean distance between each corresponding pairs of landmarks on $LF$ and $LR$, and finally normalized them with temporal length and number of landmark points. LMD is defined as:
\begin{align}
\label{eq:lmd}
LMD  =  \frac{1}{T} \times \frac{1}{P}\sum_{t=1}^{T}\sum_{p=1}^{P} {\norm{LR_{{t,p}} - {LF_{{t,p}}}}_2} 
\enspace,
\end{align}
where $T$ denotes the temporal length of video and $P$ denotes the total number of landmark points on each image (20 points).

\subsection{Ablation Study} 
\label{ssec:ablation}
\begin{table}[t]
\centering

\begin{tabular}{c|c:c:c:c:c:c:c:c:c}
\toprule
\textbf{Methods} & (a) & (b) & (c) & (d) & (e) & (f) & (g) & (h) & (i) \\ \hline
$\ell_{corr}$ &  &  &  &  & \checkmark &  &\checkmark  &  & \checkmark \\ 
$\ell_{corr}$(Non-Derivative Corr.) &  &  &  &  &  &\checkmark  &  &  &  \\ 

$\ell_{gen}$ & \checkmark  &  & \checkmark & \checkmark &\checkmark  &\checkmark  & \checkmark & \checkmark & \checkmark \\
$\ell_{pix}$ &  &\checkmark  &\checkmark  &\checkmark  &\checkmark  &\checkmark  & \checkmark & \checkmark & \checkmark \\
$\ell_{perc}$ &  &  &  & \checkmark & \checkmark &  \checkmark& \checkmark &\checkmark  & \checkmark \\ 
Two-Stream D. &  &  &  &  &  &  & \checkmark & \checkmark &  \\
Three-Stream D. &\checkmark  &  & \checkmark &\checkmark  & \checkmark &\checkmark  &  &  &  \\
\begin{tabular}[c]{@{}c@{}}Three-Stream D.(Frame-Diff.)\end{tabular} &  &  &  &  &  &  &  &  &\checkmark  \\ \hline

\multicolumn{1}{c}{\textbf{Metrics}} & \multicolumn{9}{c}{ } \\
\hline
LMD &  1.37  & 1.28  & 1.33 & 1.31 & \textbf{1.18} & 1.96 & 1.39 & 1.42 & 1.40 \\
SSIM & 0.67  & \textbf{0.79} & 0.66 & 0.70 & 0.73 & 0.52 & 0.68 & 0.59 & 0.66 \\
PSNR & 29.46 & 29.81 & 29.66 & 29.40 & \textbf{29.98} & 28.6 & 29.59 &  29.46 & 29.51  \\
CPBD & 0.176 & 0.006 & 0.182 & 0.209 & 0.175 & \textbf{0.218} & 0.187 & 0.176 & 0.210 \\ 
\bottomrule

\end{tabular}
\vspace{2mm}

\caption{Ablation results on GRID dataset. The full model (method (e)) uses all four losses as described in Sec.\ref{sec:fullmodal}. For LMD, the lower the better. SSIM, PSNR and CPBD, the higher the better. We bold each leading score.}
\label{tab:result}
\vspace{-8mm}
\end{table}









We conduct ablation experiments to study the contributions of the three components in our full model separately: correlation loss, three-stream GAN discriminator and perceptual loss. The ablation study is conducted on GRID dataset. Results are shown in Tab.~\ref{tab:result}. Different implementations are discussed in below as well.  The following ablation studies are trained and tested on the GRID dataset.

\noindent \textbf{Perceptual Loss.} \quad Generally, we find that perceptual loss can improve the result in metrics such as LMD, SSIM and CPBD, which means that perceptual loss can help our model generate more accurate lip movements with higher image quality, and improve image sharpness at the same time (see method (c) v.s. method (d) in Tab.\ref{tab:result}).

\noindent \textbf{Correlation Models.} \quad 
When correlation loss is removed from final objective function Eq.~\ref{eq:full_loss}, results are worse than final objective in LMD, SSIM and PSNR, demonstrating the importance of correlation loss in generating more accurate lip movement (see method (d) v.s. method (e) or method (g) v.s. method (h)).

Besides, we investigate a model variant, \emph{Non-Derivative Correlation} (see method (f) in Tab.~\ref{tab:result}), for analyzing the necessity of applying derivative features to $\phi_s$ and $\phi_v$. Instead of using the derivative of audio features and the optical flow, this variant just uses audio features $f_s$ and video frames $v$ directly as inputs. Neither the derivative nor the optical flow is calculated here. Other settings (e.g., network structure and loss functions) are identical with the full model (denoted as method (e) in Tab.~\ref{tab:result}). The comparison between method (e) and method (f) in Tab.~\ref{tab:result} shows that derivative correlation model outperforms the \emph{Non-Derivative Correlation} model in metrics such as SSIM, PSNR and LMD. With respect to \emph{Non-Derivative Correlation} model, landmark distance is even worse than
model without correlation loss (method (d)). The experimental result proves our assumption that it is the derivatives of audio and visual information rather than the direct features that are correlated. Furthermore, since \emph{Non-Derivative Correlation} model fails to learn the derivative feature implicitly (i.e. convolutional layers fails to transform feature to their derivatives), using the derivatives of audio and visual features to do correlation as a strong expert prior knowledge is necessary.



\noindent \textbf{GAN Discriminator.} \quad
We find that $\ell_{gen}$ improves the CPBD result (see method (a), (b) and (c) in Tab.~\ref{tab:result}), demonstrating that discriminator can improve the frame sharpness.

Furthermore, we use two model variants to study the effectiveness of proposed three-stream GAN discriminator.  
\emph{Two-Stream Discriminator} (see Two-Stream D. in Tab.~\ref{tab:result}) only contains audio stream and video stream. The \emph{Frame-Difference Three-Stream Discriminator} (see Three-Stream D.(Frame-Diff.) in Tab~\ref{tab:result}) replaces the optical flow with frame-wise difference, i.e., $L1$ distance between adjacent frames, as the third stream to capture motion changes. First, compared with the \emph{Two-Stream Discriminator} variant, our full model with proposed \emph{Three-Stream Discriminator} gives better result (see method (e) v.s. method (g)), which indicates the effectiveness of explicitly modeling motion changes among the frames. Second, compared with the \emph{Frame-Difference Three-Stream Discriminator} variant, the full model generates more realistic (higher CPBD) and accurate lip movements (lower LMD) (see method (e) and (i)), which indicates that optical flow is a better representation than frame-wise difference for modeling motion changes.


\subsection{Comparison to State-of-the-Art}
\label{sec:exp:soa}
\begin{figure}[t]
\begin{center}
\includegraphics[width=0.9 \linewidth]{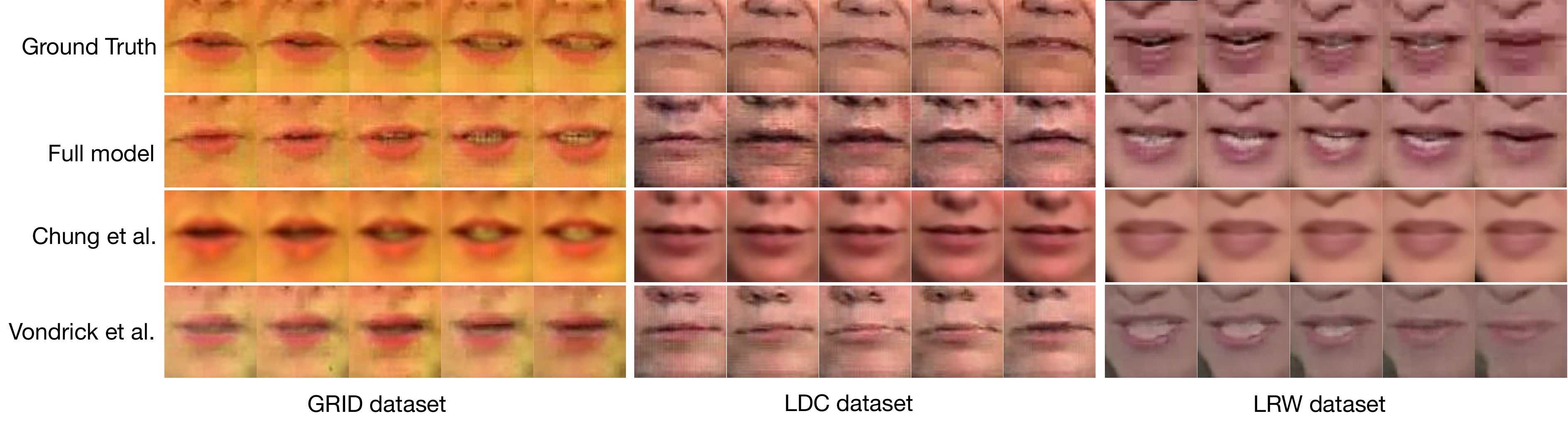}
\end{center}
\vspace{-3mm}
   \caption{Generated videos of our model on three testing datasets compared with state-of-the-art methods. In the testing set, none of the speakers were shown in the training set.}
\label{fig:three_results}
\vspace{-3mm}
\end{figure}

\begin{table*}[!htbp]
    \centering
  \begin{tabular*}{0.98\linewidth}{  p{1.62cm} p{0.7cm} p{0.75cm} p{0.75cm} p{0.85cm}|p{0.7cm} p{0.75cm} p{0.75cm} p{0.85cm}|p{0.7cm} p{0.75cm} p{0.75cm} p{0.75cm}  }
      \toprule
      \toprule
Method & \multicolumn{4}{c}{GRID} & \multicolumn{4}{c}{LDC} & \multicolumn{4}{c}{LRW}   \\
      \midrule
& { \scriptsize{LMD}} &{ \scriptsize SSIM}&{\scriptsize PSNR }& {\scriptsize CPBD }& { \scriptsize{LMD}} &{ \scriptsize SSIM}&{\scriptsize PSNR }& {\scriptsize CPBD }  & { \scriptsize{LMD}} &{ \scriptsize SSIM}&{\scriptsize PSNR}& {\scriptsize CPBD}     \\
\hline
{\scriptsize G. T. }   &{\scriptsize 0 } &{\scriptsize N/A}  &{\scriptsize N/A} &{\scriptsize 0.141} &{\scriptsize 0} &{\scriptsize N/A} &{\scriptsize N/A} &{\scriptsize 0.211}&{\scriptsize 0} &{\scriptsize N/A} &{\scriptsize N/A}&{\scriptsize 0.068} \\
 \hline

{\scriptsize Vondrick
\cite{DBLP:journals/corr/VondrickPT16} }&{\scriptsize 2.38} &{\scriptsize 0.60}   &{\scriptsize 28.45} &{\scriptsize 0.129} &{\scriptsize 2.34} &{\bf{\scriptsize 0.75}} &{\scriptsize 27.96} &{\scriptsize 0.160} &{\scriptsize 3.28} &{\scriptsize 0.34} &{\scriptsize 28.03}  &{\scriptsize 0.082} \\

 \hline
 {\scriptsize Chung
 ~\cite{DBLP:journals/corr/ChungJZ17}} &{\scriptsize 1.35}& {\bf{\scriptsize 0.74}}& {\scriptsize 29.36} &{\scriptsize 0.016}  &{\scriptsize 2.13} &{\scriptsize 0.50} &{\scriptsize 28.22} &{\scriptsize 0.010} &{\scriptsize 2.25} &{\scriptsize 0.46} &{\scriptsize 28.06} &{\bf{\scriptsize 0.083}}   \\
   \hline
 {\scriptsize Full model }  & {\bf{\scriptsize 1.18}}     & {\scriptsize 0.73 }  & {\bf{\scriptsize  29.89}}    &{\bf{\scriptsize 0.175}} & {\bf{\scriptsize 1.82}}   &{\scriptsize 0.57 } & { \bf{\scriptsize 28.87}} & {\bf{\scriptsize 0.172}} &{\bf{\scriptsize1.92}} &{\scriptsize \textbf{0.53}} &{\scriptsize \bf{28.65}} & {\scriptsize 0.075}\\

      \bottomrule
  \end{tabular*}
  \vspace{2mm}
  \caption{Results on three datasets. Models mentioned in this table are trained from scratch (no pre-training included) and be tested on each dataset a time. We bold each leading score.}
    \label{tab:main_tb}
    \vspace{-5mm}
\end{table*}

In this section, we compare our full model with two state-of-the-art methods~\cite{DBLP:journals/corr/VondrickPT16,DBLP:journals/corr/ChungJZ17}. We extend~\cite{DBLP:journals/corr/VondrickPT16} to a conditional GAN structure, which receives the same target image information and audio information as our models. 
The quantitative results are shown in Tab.~\ref{tab:main_tb}. We test our models on three different datasets. The results show that our proposed models outperform state-of-the-art models in most of the metrics. In terms of LMD and PSNR, our full model shows better performance than methods that use discriminator~\cite{DBLP:journals/corr/VondrickPT16} or reconstruction loss~\cite{DBLP:journals/corr/ChungJZ17}. Model proposed by Chung et al., based on reconstruction loss, generates much more blurred images, which makes them look unrealistic. We can see this phenomenon in the CPBD column. The LRW dataset consists of people talking in the wild so resolution is much smaller in terms of lip region. We need to scale up the ground truth to 64 $\times$ 64, which leads to a lower resolution and CPBD. We suspect this is the reason why we achieve a better CPBD than ground truth in LRW dataset.

\begin{figure}[t]
\centering
\includegraphics[width=0.9\linewidth]{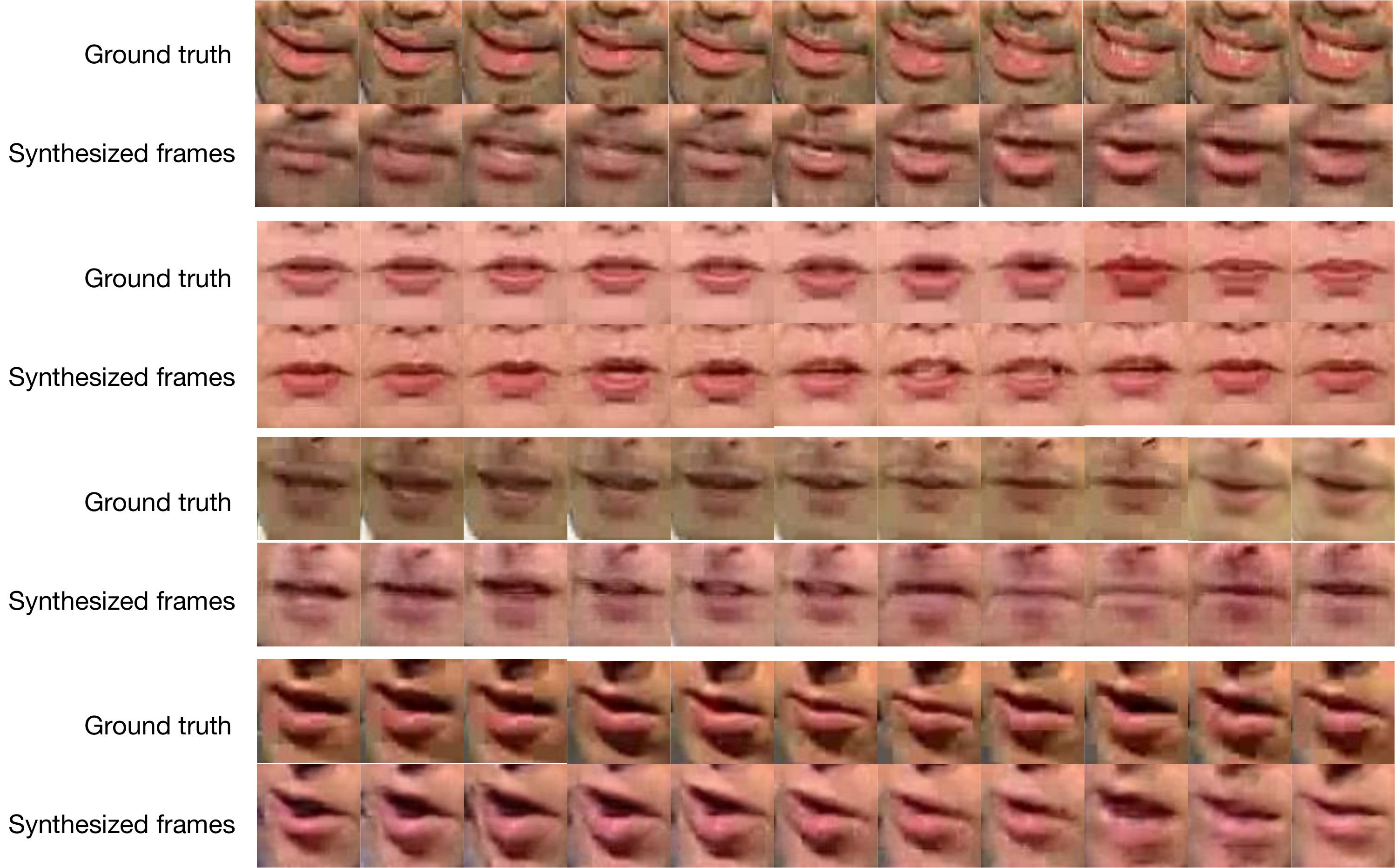}
\caption{\label{fig:lrw_result} Randomly selected outputs of the full model on the LRW testing set. The lip shape in videos not only synchronize well with the ground truth, but maintain identity information, such as (beard v.s. no beard).}
\vspace{-3mm}
\end{figure}
The qualitative results compared with other methods are shown in Fig.~\ref{fig:three_results}. Our model generates sharper video frames on all three datasets, which has also been supported by the CPBD results, even if input identity images are in low resolution. We show additional results of our method in Fig.~\ref{fig:lrw_result}. Our model can generate realistic lip movement videos that are robust to view angles, lip shapes and facial characteristics in most of the times. However, sometimes our model fails to preserve the skin color (see the last two examples in Fig.~\ref{fig:lrw_result}), which, we suspect, is due to the imbalanced data distribution in LRW dataset. Furthermore, the model has difficulties in capturing the amount of lip deformations of each person, which is an intrinsic problem when learning from a single image.





\subsection{Real-World Novel Examples}
\label{sec:exp:novel}

\begin{figure*}[ht]
\centering
\includegraphics[width=0.9\linewidth]{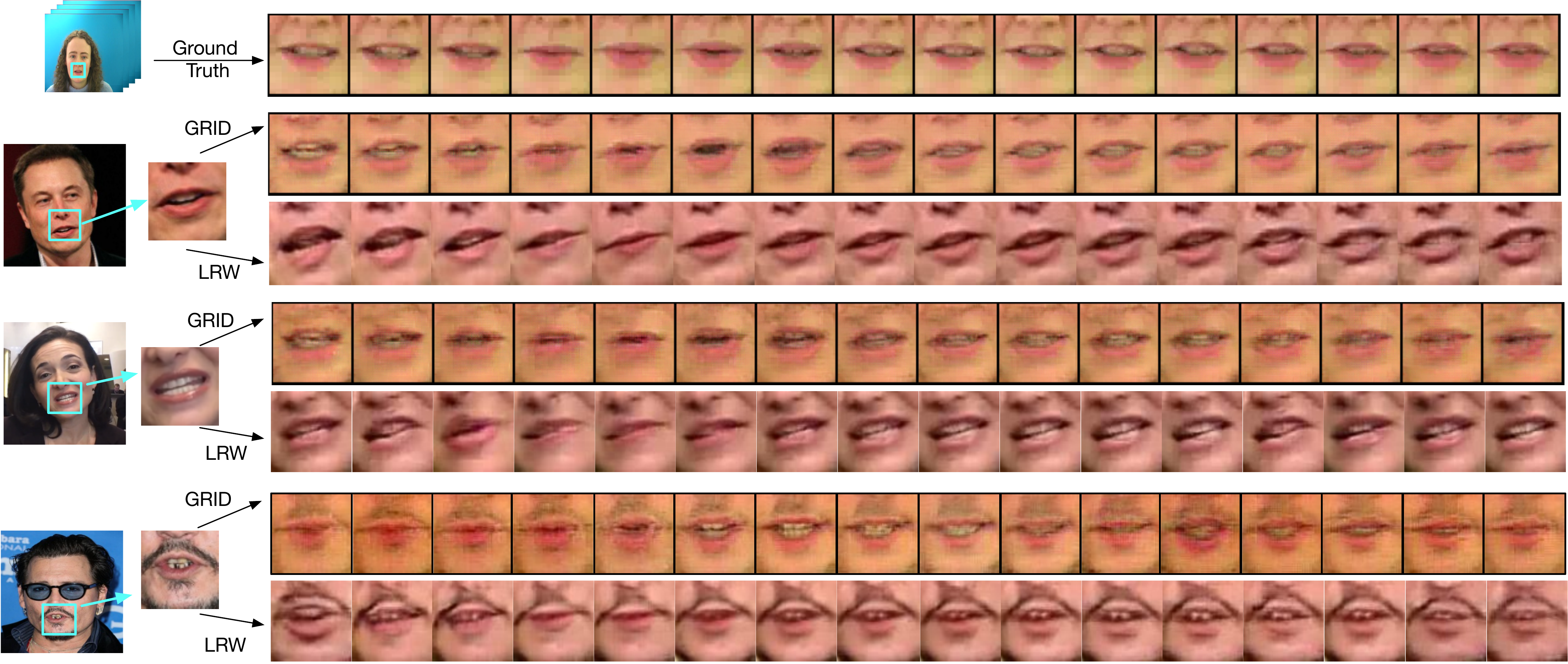}
\caption{\label{fig:new_identity}
The figure shows the generated images based on three identity images outside of dataset, which is also not paired with the input audio from GRID dataset. Two full models trained on GRID and LRW datasets are used here for a comparison.
}
\vspace{-3mm}
\end{figure*}


For generating videos given unpaired identity image and audio in the real-world, i.e., source identity of provided audio is different from the target identity and out of the datasets, our model can still perform well. Results are shown in Fig.~\ref{fig:new_identity}, in which three identity images of celebrities are selected outside of the datasets the model trained on and the input audio is selected in GRID dataset. For our model trained on LRW, both identity images and audio are unseen. For our model trained on GRID, we leave the source identity out of the training.

The videos generated by our model show promising qualitative performance. Both lip regions of Musk and Sandburg are rotated by some degrees. We can see that the rotation phenomenon in the generated video frames as well. Besides, our model can also retain beards in our generated clip when identity (target person) has beards as well. However, we observe that model trained on GRID dataset fails to reserve the identity information. Because of the fact that LRW dataset has much more identities than GRID dataset (hundreds v.s. 33), the model trained on LRW has better generalization ability.


\section{Conclusion and Future Work}
\label{sec:conclusion}

In this paper, we study the task: given an arbitrary audio speech and one lip image of arbitrary target identity, generate synthesized lip movements of the target identity saying the speech. To perform well in this task, it requires a model to not only consider the retention of the target identity, photo-realistic of synthesized images, consistency and smoothness of images in a video, but also learn the correlations between the speech audio and lip movements. We achieve this by proposing a new generator network, a novel audio-visual correlation loss and a full model that considers four complementary losses. We show significant improvements on three datasets compared to two state-of-the-art methods. There are several future directions. First, non-fixed length lip movements generation is needed for a more practical purpose. Second, it is valuable to extend our method to one generating full face in an end-to-end paradigm.\\

\noindent {\small \textbf{Acknowledgement} \quad This work was supported in part by NSF BIGDATA 1741472 and the University of Rochester AR/VR Pilot Award. We gratefully acknowledge the gift donations of Markable, Inc., Tencent and the support of NVIDIA with the donation of GPUs used for this research. This article solely reflects the opinions and conclusions of its authors and not the funding agents.}

{\small
\bibliographystyle{splncs}
\bibliography{egbib}
}

\clearpage

  \begin{subappendices}
\renewcommand{\thesection}{\Alph{section}}%

\section{Network Architectures}
\label{sec:net}
This section details the network structures mentioned in the submitted paper, including \emph{Audio Encoder} (see Tab.~\ref{tab:AudioEncoder}), \emph{Identity Encoder} (see Tab.~\ref{tab:IdentityEncoder}), \emph{Decoder} (see Tab.~\ref{tab:Decoder}), \emph{Audio Derivative Encoder} (see Tab.~\ref{tab:AudioDerivativeEncoder}), \emph{Flow Encoder} (see Tab.~\ref{tab:FlowEncoder}), and \emph{Three-stream GAN discriminator} (see Tab.~\ref{tab:DiscAudio}, Tab.~\ref{tab:DiscVideo} and Tab.~\ref{tab:DiscOpt}). For simplicity, all "Conv"s in the following tables stand for the sequence of Convolution, Batch Normalization and ReLU.

\begin{table}[ht!]
\centering
\begin{tabular}{l|l|l|l|l}
\toprule
\hline
\textbf{Layers} & \textbf{Output Size} & \textbf{Kernel} & \textbf{Stride} & \textbf{Padding} \\ \hline
Conv            & $16 \times 128$     & $3 \times 3$    & 1, 1            & 1, 1             \\ \hline
Conv            & $16 \times 64$      & $3 \times 3$    & 1, 2            & 1, 1             \\ \hline
Conv            & $16 \times 64$      & $3 \times 3$    & 1, 1            & 1, 1             \\ \hline
Conv            & $16 \times 32$      & $3 \times 3$    & 1, 2            & 1, 1             \\ \hline
Max Pooling     & $16 \times 16$      & $1 \times 2$    & 1, 2            & 0, 0             \\ \hline
\end{tabular}
\caption{Network Structure of Audio Encoder.}
\label{tab:AudioEncoder}
\end{table}

\begin{table}[ht!]
\centering
\begin{tabular}{l|l|l|l|l}
\toprule
\hline
\textbf{Layers} & \textbf{Output Size} & \textbf{Kernel} & \textbf{Stride} & \textbf{Padding} \\ \hline
Conv            & $64 \times 64$      & $7 \times 7$    & 1, 1            & 3, 3             \\ \hline
Conv            & $32 \times 32$      & $3 \times 3$    & 2, 2            & 1, 1             \\ \hline
Conv            & $16 \times 16$      & $3 \times 3$    & 2, 2            & 1, 1             \\ \hline
\end{tabular}
\caption{Network Structure of Identity Encoder.}
\label{tab:IdentityEncoder}
\end{table}

\begin{table}[ht!]
\centering
\begin{tabular}{l|l|l|l|l|l}
\toprule
\hline
\textbf{Layers} & \textbf{Output Size}         & \textbf{Kernel}                                                            & \textbf{Stride}                                             & \textbf{Padding}                                            & \textbf{\thead{Output \\ Padding}}                       \\ \hline
3D ResBlocks      & $  16 \times 16 \times 16$ & $\begin{bmatrix}1\times3\times3 \\ 1\times 3 \times 3\end{bmatrix} \times 9$ & $\begin{bmatrix}(1, 1, 1) \\ (1, 1, 1)\end{bmatrix} \times 9$ & $\begin{bmatrix}(0, 1, 1) \\ (0, 1, 1)\end{bmatrix} \times 9$ & -                                               \\ \hline
Trans. Convs  & $16\times 64 \times 64$ & $\begin{bmatrix}3\times3\times3 \end{bmatrix} \times 2$                      & $\begin{bmatrix}1, 2, 2 \end{bmatrix} \times 2$               & $\begin{bmatrix}1, 1, 1 \end{bmatrix} \times 2$               & $\begin{bmatrix}0, 1, 1 \end{bmatrix} \times 2$ \\ \hline
Convolution       & $16 \times 64 \times 64$ & $7 \times 7 \times 7$                                                        & 1, 1, 1                                                       & 3, 3, 3                                                       & -                                  \\ \hline
Tanh              & $16 \times 64 \times 64$ & -                                                              & -                                                & -                                                & -                                 \\ \hline
\end{tabular}
\caption{Network Structure of Decoder. One "Trans. Convs" stands for a sequence of Transposed Convolution, Batch Normalization and ReLU.}
\label{tab:Decoder}

\end{table}

\begin{table}[t]
\centering
\begin{tabular}{l|l|l|l|l}
\toprule
\hline
\textbf{Layers} & \textbf{Output Size} & \textbf{Kernel} & \textbf{Stride} & \textbf{Padding} \\ \hline
Conv            & $16 \times 8$       & $3 \times 3$    & 1, 2            & 1, 1             \\ \hline
Conv            & $16 \times 4$       & $3 \times 3$    & 1, 2            & 1, 1             \\ \hline
Conv            & $16 \times 2$       & $3 \times 3$    & 1, 2            & 1, 1             \\ \hline
Conv            & $16 \times 1$       & $3 \times 3$    & 1, 2            & 1, 1             \\ \hline
\end{tabular}
\caption{Network Structure of Audio Derivative Encoder (denoted as $\phi_s$ in the main paper).}
\label{tab:AudioDerivativeEncoder}

\end{table}

\begin{table}[t]
\centering
\begin{tabular}{|l|l|l|l|l|}
\hline
\textbf{Layers} & \textbf{Output Size}    & \textbf{Kernel}       & \textbf{Stride} & \textbf{Padding} \\ \hline
Conv            & $16 \times 8 \times 8$ & $3 \times 3 \times 3$ & 1, 2, 2         & 1, 1             \\ \hline
Conv            & $16 \times 4 \times 4$ & $3 \times 3 \times 3$ & 1, 2, 2         & 1, 1             \\ \hline
Conv            & $16 \times 2 \times 2$ & $3 \times 3 \times 3$ & 1, 2, 2         & 1, 1             \\ \hline
Conv            & $16 \times 1 \times 1$ & $3 \times 3 \times 3$ & 1, 2, 2         & 1, 1             \\ \hline
\end{tabular}
\caption{Network Structure of Flow Encoder (denoted as $\phi_v$ in the main paper).}
\label{tab:FlowEncoder}

\end{table}

\begin{table}[t]
\centering
\begin{tabular}{l|l|l|l|l}
\toprule
\hline
\textbf{Layers} & \textbf{Output Size} & \textbf{Kernel} & \textbf{Stride} & \textbf{Padding} \\ \hline
Conv            & $16 \times 128$     & $3 \times 3$    & 1, 1            & 1, 1             \\ \hline
Conv            & $16 \times 64$      & $3 \times 3$    & 2, 2            & 1, 1             \\ \hline
Conv            & $16 \times 64$      & $3 \times 3$    & 1, 1            & 1, 1             \\ \hline
Conv            & $16 \times 32$      & $3 \times 3$    & 2, 2            & 1, 1             \\ \hline
FC     & $256$    & -            & -   &-         \\ \hline
\end{tabular}
\caption{Network Structure of audio stream in Three-stream GAN discriminator. FC stands for fully connected layer.}
\label{tab:DiscAudio}

\end{table}

\begin{table}[t]
\centering
\begin{tabular}{l|l|l|l|l}
\toprule
\hline
\textbf{Layers} & \textbf{Output Size}    & \textbf{Kernel}       & \textbf{Stride} & \textbf{Padding} \\ \hline
Conv            & $8 \times 32 \times 32$ & $4 \times 4 \times 4$ & 2, 2, 2         & 1, 1             \\ \hline
Conv            & $4 \times 16 \times 16$ & $4 \times 4 \times 4$ & 2, 2, 2         & 1, 1             \\ \hline
Conv            & $ 2 \times 8 \times 8$ & $4 \times 4 \times 4$ & 2, 2, 2         & 1, 1             \\ \hline
Conv            & $1 \times 4 \times 4$ & $ 4 \times 4 \times 4$ & 1, 2, 2         & 1, 1             \\ \hline
\end{tabular}
\caption{Network Structure of video stream in Three-stream GAN discriminator.}
\label{tab:DiscVideo}

\end{table}

\begin{table}[t]
\centering
\begin{tabular}{l|l|l|l|l}
\toprule
\hline
\textbf{Layers} & \textbf{Output Size}    & \textbf{Kernel}       & \textbf{Stride} & \textbf{Padding} \\ \hline
Conv            & $8 \times 8 \times 8$ & $3 \times 3 \times 3$ & 2, 2, 2         & 1, 1             \\ \hline
Conv            & $4 \times 8 \times 8$ & $3 \times 3 \times 3$ & 2, 2, 1         & 1, 1             \\ \hline
Conv            & $2 \times 4 \times 4$ & $3 \times 3 \times 3$ & 2, 2, 2         & 1, 1             \\ \hline
Conv            & $1 \times 4 \times 4$ & $3 \times 3 \times 3$ & 2, 2, 1         & 1, 1             \\ \hline
\end{tabular}
\caption{Network Structure of optical flow stream in Three-stream GAN discriminator.}
\label{tab:DiscOpt}

\end{table}

\end{subappendices}
\end{document}